\documentclass[letterpaper, 10 pt, conference]{ieeeconf}  

\IEEEoverridecommandlockouts                              

\usepackage{graphicx} 
\usepackage{subfigure}	
\usepackage[table]{xcolor}
\usepackage{diagbox}
\usepackage{multirow}
\usepackage{amsmath} 
\usepackage{amsfonts,amssymb}
\usepackage{color}
\usepackage{cite}
\usepackage{booktabs}
\usepackage{pifont}
\usepackage[capitalize]{cleveref}

\title{\LARGE \bf
BPT: Binary Point Cloud Transformer for Place Recognition
}

\author{Zhixing Hou$^{1}$, Yuzhang Shang$^{2}$, Tian Gao$^{1}$, Yan Yan$^{2}$ 
\thanks{$^{1}$ Zhixing Hou and Tian Gao are with the School of Computer Science and Engineering, Nanjing University of Science and Technology, Nanjing, China. {\tt\small E-mail: hzx@njust.edu.cn, gaotian970228@njust.edu.cn}}
\thanks{$^{2}$ Yuzhang Shang and Yan Yan are with the Computer Science Department, Illinois Institute of Technology, Chicago, USA. {\tt\small E-mail: yyan34@iit.edu, yshang4@hawk.iit.edu}}
}

\begin{document}

\maketitle

	
    \begin{abstract}
        Place recognition, an algorithm to recognize the re-visited places, plays the role of back-end optimization trigger in a full SLAM system. Many works equipped with deep learning tools, such as MLP, CNN, and transformer, have achieved great improvements in this research field. Point cloud transformer is one of the excellent frameworks for place recognition applied in robotics, but with large memory consumption and expensive computation, it is adverse to widely deploy the various point cloud transformer networks in mobile or embedded devices. To solve this issue, we propose a binary point cloud transformer for place recognition. 
        As a result, a 32-bit full-precision model can be reduced to a 1-bit model with less memory occupation and faster binarized bitwise operations. To our best knowledge, this is the first binary point cloud transformer that can be deployed on mobile devices for online applications such as place recognition.
        Experiments on several standard benchmarks demonstrate that the proposed method can get comparable results with the corresponding full-precision transformer model and even outperform some full-precision deep learning methods. For example, the proposed method achieves 93.28\% at the top @1\% and 85.74\% at the top @1\% on the Oxford RobotCar dataset in terms of the metric of the average recall rate. Meanwhile, the size and floating point operations of the model with the same transformer structure reduce  56.1\% and 34.1\% respectively from original precision to binary precision.
    \end{abstract}
    
    
    \section{Introduction}
    The Simultaneous Localization and Mapping (SLAM) algorithm, an important prerequisite for robot navigation, plays a key role in robotics and autonomous driving research. Accumulated pose error drift over time due to intrinsic sensor noise seriously affects the mapping quality in the long-range SLAM system. 
    Place verification with the point cloud (i.e., loop closure detection in robotics society) means a robot equipped with the LIDAR sensor should able to 
    recognize the places that have been visited before. The exact place recognition can trigger the back-end optimization of the SLAM system to reduce the accumulated pose error and map distortion.

    Many promising point-cloud based place recognition algorithms \cite{Wang2020LiDARIF,kim2018scan,segmatch2017, Uy2018PointNetVLADDP, Liu-LPD-Net, du2020dh3d, vid2021locus, Xia2020SOENetAS, Zhou2021NDTTransformerL3, hou2022hitpr, lcdnet, Komorowski_2021_WACV, chen2020overlapnet, he2016m2dp} have been proposed so far. Among them, some use classical feature extraction methods, such as \cite{Wang2020LiDARIF, kim2018scan}, with the help of well-designed parameters and data organization format, resulting in fast execution and good recognition accuracy but being sensitive to variation of the environment over a long period. Complementarily, other methods \cite{Uy2018PointNetVLADDP, Liu-LPD-Net, du2020dh3d, vid2021locus, Xia2020SOENetAS, Zhou2021NDTTransformerL3, hou2022hitpr,lcdnet, Komorowski_2021_WACV} based on deep learning frameworks are usually robust to the variant environment of the re-visited places but consume much more computation time and memory. 	
    Among the deep learning based methods, the transformer-based ones are the focus of our attention due to their outstanding performance in the place recognition task. With the power of the self-attention mechanism, the transformer models have already made significant achievements in natural language processing \cite{devlin2019bert, brown2020language}, image processing \cite{dosovitskiy2020vit, liu2021swin} and point cloud processing \cite{zhao2021point, guo2021pct}, especially in the large scale pre-trained tasks. In terms of specific tasks, several works have made some attempts with the transformer on the place recognition task and achieved good results including SOE-Net \cite{Xia2020SOENetAS}, NDT-Transformer \cite{Zhou2021NDTTransformerL3}, HiTPR \cite{hou2022hitpr}, PPT-Net \cite{hui2021pyramid} and SVT-Net \cite{fan2022svt}.
    
    However, a serious problem when applying these transformer models in the real world is that transformer models generally depend on heavy computation resources. As a result, it is adverse to the intention of their widespread deployment on mobile or wearable devices.   
    To deal with the limitations of transformer models in real-time applications, model compression strategies have been proposed to reduce the number of parameters while still preserving comparable performances. These methods include but are not limited to pruning~\cite{fan2019reducing,michel2019sixteen,voita2019analyzing}, knowledge distillation~\cite{jiao2020tinybert,sun2019patient,sanh2019distilbert}, low-rank approximation~\cite{ma2019tensorized,lan2019albert} and adaptive dynamic networks~\cite{hou2020dynabert,liu2020fastbert}. Compared with the above methods, model compression through quantitation ~\cite{zafrir2019q8bert,zhang2020ternarybert,shen2020q,lizhexin2022qvit,liyanjing2022qvit}  accelerates the execution efficiency of the algorithm without adjusting the overall network structure, reduces the dependence on high-performance devices, and is conducive to the application of general network in specific practical problems without the need of redesigning the network structure and optimization strategy for specific problems.
    Among these quantization methods, the extreme quantization methods~\cite{bai2021binarybert, qin2021bibert, liu2022bit} push the limits of the transformer compression to 1-bit binary values from the original 32-bit full-precision ones. In the binary quantization methods, the model size is reduced substantially and the calculation speed is also accelerated rapidly by using the binary bitwise operations~\cite{rastegari2016xnor} including $xnor$ and $bitcount$, which is the significant advantage over the other non-binary quantization methods. 

    However, generalizing prior quantization methods into cloud transformers is non-trivial, as unlike natural language and image, due to the irregularity of point cloud representation, the point cloud transformer model~\cite{zhao2021point,guo2021pct, wu2022point} does not have a unified standard paradigm.
    Therefore, the binarization of a point cloud transformer usually requires a redesigned solution. To our best knowledge, no binary point cloud transformer model has been proposed so far, and we propose and implement the first \textbf{B}inary \textbf{P}oint cloud \textbf{T}ransformer network called \textbf{BPT} based on the full-precision counterpart~\cite{guo2021pct}. 
    
    To verify the effectiveness of our scheme, we first apply the BPT model to a simple point cloud classification problem. Experiments on the ModelNet40 dataset \cite{wu20153d} \textbf{get 93.43\% overall accuracy, which even outperforms the full-precision point cloud transformer}. Furthermore, we apply the BPT model to the point cloud place recognition task and evaluate the performance on several standard benchmarks. 
    Our model \textbf{achieves 93.28\% at the top @1\% and 85.74\% at the top @1} on the Oxford RobotCar dataset in terms of the average recall rate metric, with almost no performance drops compared with the full-precision counterpart while \underline{\textbf{reducing 56.1\% model size and 34.1\% FLOPs}}. It demonstrates the great advantages and potential of applying the BPT model in the real-world point cloud place recognition tasks on resource-constrained devices.
    
        
        


    \section{Related works}
    \label{suc:related_works}

    In this section, we show some works closely related to ours, including works for place recognition and binary quantization, and introduce details of the basic full-precision point cloud transformer model on which our work is based.

    \subsection{Place Recognition}
        \noindent\textbf{Handcrafted descriptors:}
        At the earliest, descriptors used for place recognition, such as ESF (Ensemble of Shape Functions) \cite{ESF}, PFH (Persistent Feature Histogram) \cite{point-hist} and FPFH (Fast Point Feature Histogram) \cite{fpfh}, are well-designed features through a series of artificial extractors. Segmentation-based algorithm SegMatch \cite{segmatch2017} roughly clustered 3D shapes in the environment and matches the correspondences of these shapes between two point clouds to recognize the revisited places. M2dp \cite{he2016m2dp} built the descriptor from the density signature of multiple 2D planes projected from the 3D point cloud. In the expanded bird's view, LiDAR-Iris \cite{Wang2020LiDARIF} and Scan-Context \cite{kim2018scan} encoded height information of objects around the robot itself to construct a global descriptor.

	\noindent\textbf{Learning-based descriptors:}
        As the first proposed end-to-end method for point cloud place recognition, PointNetVLAD \cite{Uy2018PointNetVLADDP} created a precedent and paradigm with the deep learning framework by combining the work for point cloud processing named PointNet \cite{pointnet} and the one used in the field of image recognition named NetVLAD \cite{NetVlad}. The former is used as a local feature extractor and the latter is used as a pooling layer to aggregate the point-wise features. 
        Subsequently, LPD-Net \cite{Liu-LPD-Net} extracted learning-based point-wise features by adding a small graph neural network proposed in DGCNN \cite{Wang2019DynamicGC} to the foregoing work PointNetVLAD \cite{Uy2018PointNetVLADDP} and concatenated the learning-based features with the additional handcrafted features, such as change of curvature and 2D scattering, to enhance the local information. PCAN \cite{Zhang2019PCAN3A} utilized PointNet++ \cite{Qi2017PointNetDH} to convert the local point-wise features to an attention map, which was then fed into the NetVLAD \cite{NetVlad} layer during the aggregation procedure. EPC-Net \cite{Hui2021Efficient3P} introduced the spatial adjacent matrix and proxy points to compute edge convolution with lower memory consumption. In addition, it improved NetVLAD with a Grouped VLAD which contributed to reducing the number of parameters of the network.
        DH3D \cite{du2020dh3d}, LCDNet \cite{lcdnet} and OverlapNet \cite{chen2020overlapnet} fed two frames of the point cloud data into the same shared encoder structure, then recognized the revisited places and estimated the relative pose transformations simultaneously. 
        From the perspective of the transformer, NDT-Transformer \cite{Zhou2021NDTTransformerL3} and SOE-Net \cite{Xia2020SOENetAS} used the handcrafted NDT cells and raw point data as the input of the network, respectively, followed by feature-wise transformer to build the global relations. As a pure transformer network, HiTPR \cite{hou2022hitpr} combined the short-range and long-range transformer structure to extract the local feature and global contextual relations to build the point cloud descriptor. 
        However, compared with our work, those transformer-based networks are not conducive to board deployment on mobile devices, due to their relatively heavy resource dependency.
    
    \subsection{Binary quantization}
    Binary quantization (also dubbed binarization), compressing the weights and activations of the neural network from 32-bit to 1-bit, prompts the network to possess less storage resource requirements and faster computational velocity. Courbariaux et al.\cite{courbariaux2016binarized} first proposed the concept of the binary neural network, through two kinds of state quantities +1 and -1 to fit full-precision model parameters and activation values, achieved the effect of compressing the model and improving the operation efficiency. Subsequently, Rastegari et al. \cite{rastegari2016xnor} introduced XNOR-Net, and demonstrated the effectiveness of the model through rigorous theoretical analysis and experiments on large classification datasets such as ImageNet1K. This approach can theoretically reduce the model size by $1/32$ and speed up inference by 58 times.
    
    With the discovery of the transformer structure \cite{vaswani2017attention}, the attendant problem is the expensive consumption of memory and computing time. Numerous researchers found that binary quantization also has great potential in solving this problem. Just like the domain where the transformer was first discovered, binarization for the transformer structure is also rising in the field of natural language processing. 
    First, Bai et al. \cite{bai2021binarybert} proposed a transformer binarization work named BinaryBERT by equivalently splitting from a half-size ternary BERT network \cite{zhang2020ternarybert} that pushes the transformer quantization to the limit. Qin et al. \cite{qin2021bibert} proposed BiBERT which introduces the Bi-Attention module and Direction-Matching distillation method from the perspective of information theory by maximizing the information entropy of the binarized representations. Liu et al. \cite{liu2022bit} also implemented a binarized BERT model called BiT that includes a two-set binarization scheme that utilizes different binarization methods to compact activations with and without non-linear activation layers. Furthermore, they made use of the multi-step distillation method, which first distills the full-precision model into an intermediate model and then distills the intermediate model into the binarized transformer model.
    
    As for vision transformer research fields, Li et al. \cite{lizhexin2022qvit} compacted parameters and activations of the vision transformer model to 3-bit and Li et al. \cite{liyanjing2022qvit} further pushed the limit of vision transformer quantization to 2-bit. However, there has been no binarized vision transformer, i.e. 1-bit weights and activation representations proposed so far. 
    
    In the same situation as the vision transformer, few binarization works are applied to point cloud processing tasks. The first binary neural network for raw point cloud processing proposed by Qin et al. \cite{qin2020bipointnet} introduced Entropy-Maximizing Aggregation and Layer-wise Scale Recovery to effectively prevent immense performance drop. Xu et al. \cite{xu2021poem} introduced a bi-modal distribution concept to mitigate the impact of small disturbances and exploited Expectation-Maximization to constrain the weights of the network into this distribution. Su et al. \cite{su2022svnet} designed a binarization and training method for the point cloud network with invariant scales and equivariant vectors simultaneously. But these three works are all based on the MLP structure originated from the classic work PointNet \cite{pointnet} without the participation of the transformer structure. So our work proposed in this paper is the first binary point cloud transformer model and it is also the first time to be applied to place recognition.

    \subsection{Full-precision point cloud transformer}
    \label{subsec:fp-pct}
    There are several different types of transformer-related works to deal with point cloud tasks. Among them, the most typical works are Point Transformer \cite{zhao2021point}, Point Transformer V2 \cite{wu2022point}, and PCT (Point Cloud Transformer) \cite{guo2021pct}. PCT \cite{guo2021pct}, used for point cloud classification and segmentation proposed by Guo et al., stacked multiple self-attention layers and concatenated the features from each transformer block as contextual information. Point Transformer \cite{zhao2021point} and Point Transformer V2 \cite{wu2022point} focused more on superior local features, achieving state-of-the-art performance on classification and segmentation.

\begin{figure*}[t]
        \centering
        \subfigure[]{
            \begin{minipage}[t]{0.15\textwidth}
                \centering
                \includegraphics[width=1\textwidth]{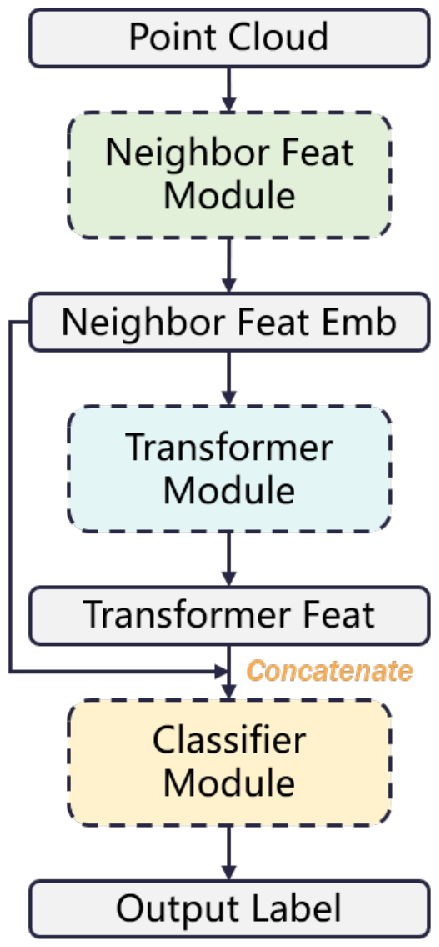}
                \label{fig:pct_a}
            \end{minipage}%
        }%
        \subfigure[]{
            \begin{minipage}[t]{0.18\textwidth}
                \centering
                \includegraphics[width=1\textwidth]{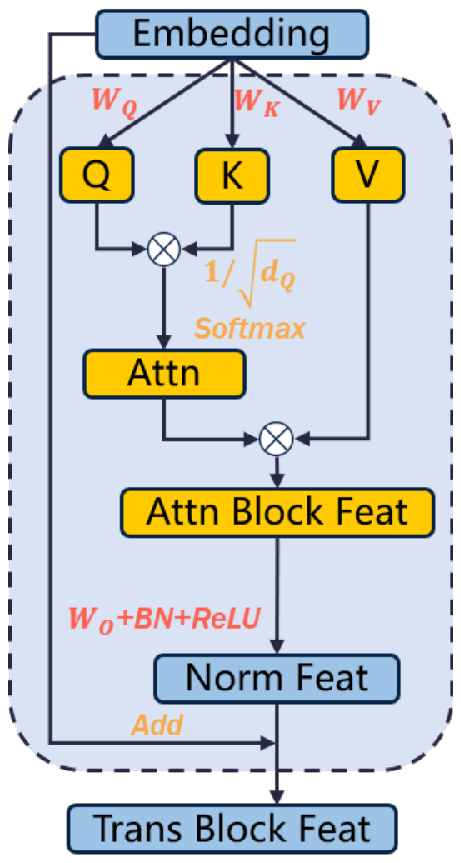}
                \label{fig:pct_b}
            \end{minipage}%
        }%
        \subfigure[]{
            \begin{minipage}[t]{0.18\textwidth}
                \centering
                \includegraphics[width=1\textwidth]{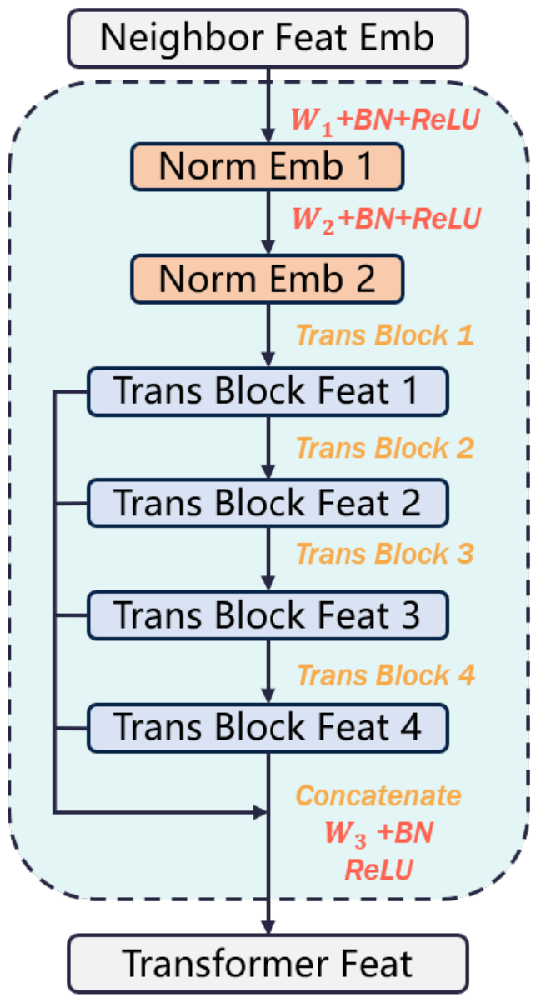}
                \label{fig:pct_c}
            \end{minipage}%
        }%
        \subfigure[]{
            \begin{minipage}[t]{0.18\textwidth}
                \centering
                \includegraphics[width=1\textwidth]{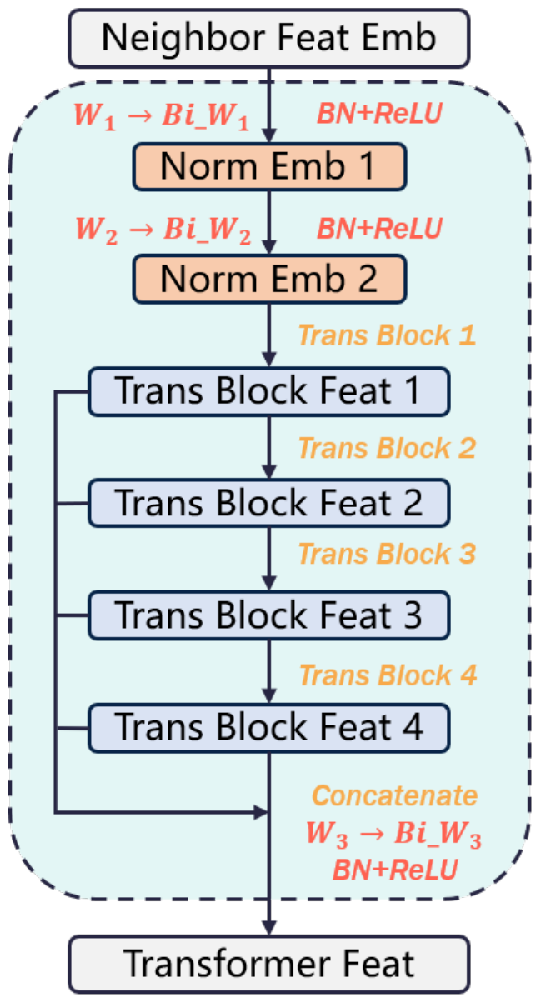}
                \label{fig:pct_d}
            \end{minipage}%
        }%
        \subfigure[]{
            \begin{minipage}[t]{0.22\textwidth}
                \centering
                \includegraphics[width=1\textwidth]{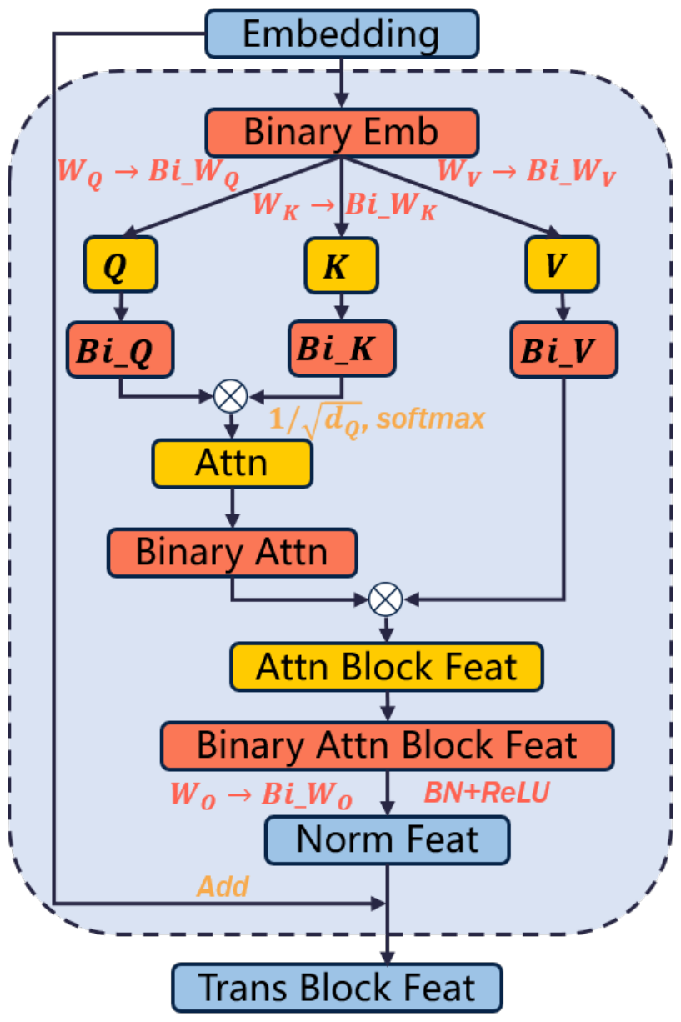}
                \label{fig:pct_e}
            \end{minipage}%
        }%
        \centering
        \caption{(a) Point cloud \textbf{transformer model}. Colored dashed boxes represent all modules contained in the model. (b) Full-precision point cloud \textbf{transformer block}. The background color indicates its position in the transformer module. (c) Full-precision point cloud \textbf{transformer module} which contains four transformer blocks. The background color indicates its position in the transformer model. (d) Binary point cloud \textbf{transformer module} which contains four binary transformer blocks. (e) binary point cloud \textbf{transformer block}. Orange words and boxes indicate the same structures both in the full-precision and binary transformer models, while Red ones indicate variations between the full-precision and binary transformer models. The symbol $\otimes$ in binary transformer blocks indicates binary operation including $XNOR$ and $Bitcount$, equivalent to dot product operation in full-precision transformer blocks. (In this paper, the transformer block is regarded as a part of the transformer module.)}
	\label{fig:transformer_model}
    \end{figure*}

    Due to its clear, concise, and effective structure shown in \cref{fig:pct_a,fig:pct_b,fig:pct_c}, the PCT model \cite{guo2021pct} is chosen as the full-precision model counterpart of our binary model. We divide the whole pipeline into three main components and name them the neighbor feature extraction module, the transformer module, and the classifier module. The neighbor feature extraction module is used to sample the central points, index the adjacent points, aggregate the local points and then generate the neighbor feature embeddings which will be fed into the transformer module in the next stage. 
    The transformer module (Fig.~\ref{fig:pct_c}) contains an input linear block ($W_1+BN+ReLU$ and $W_2+BN+ReLU$), four transformer blocks each of which is shown in Fig.~\ref{fig:pct_b}, and an output linear block ($Concatenate$, $W_3+BN$ and $ReLU$ in the Fig.~\ref{fig:pct_c}). The third part of the model is a classifier module which can also be instead by a variable task-related neural network.
    
    \begin{figure}[t]
      \centering
       \includegraphics[width=0.5\linewidth]{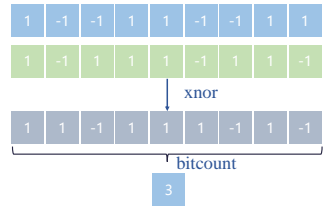}
       \caption{Bitwise binary operations: $xnor$ and $bitcount$.}
       \label{fig:bit_op}
    \end{figure}

	\section{Binary point cloud transformer}
    \label{sec:bpt}
    
    In this section, we systematically introduce the proposed \textbf{B}inary \textbf{P}oint cloud \textbf{T}ransformer (\textbf{BPT}) model based on its full-precision counterpart~\cite{guo2021pct}, including the binarization scheme of the different network layers, the variant parts of the network in the binarization procedure, and its application in the place recognition task by combining it with metric learning. 

    \subsection{Weights and activations binarization}
    \label{subsec:w_act_binarization}
    Following XNOR-Net~\cite{rastegari2016xnor}, we use the $sign()$ function as the binarization function to quantify the model weights and activations to two kinds of state quantities +1 and -1 in forward propagation, and the straight-through estimator (STE)~\cite{bengio2013estimating} in the backward propagation to update the model parameters:
    
    \begin{equation}
      Bi(x)=sign(x) = \left\{\begin{array}{ccl}
            1 & \mbox{for}
            & x\geq 0 \\ -1 & \mbox{for} & otherwise \\
                \end{array} \right.
      \label{eq:sign}
    \end{equation}
    and
    \begin{equation}
      \frac{\partial L}{\partial x} = \left\{\begin{array}{ccl}
            \frac{\partial L}{\partial sign(x)} & \mbox{for} & |x| \leq 1 \\
            0 & \mbox{for} & otherwise \\
                \end{array} \right.,
      \label{eq:ste}
    \end{equation}
    where $L$ is the loss function. Different from the activations in XNOR-Net \cite{rastegari2016xnor},  in our proposed model, the binarization function for the activations following non-linear functions (i.e. $ReLU()$ and $softmax()$) is expressed as Eq.\ref{eq:act_sign} due to their non-negative.
    \begin{equation}
      Bi(x)= \left\{\begin{array}{ccl}
            1 & \mbox{for}
            & x\geq 0.5 \\ 0 & \mbox{for} & otherwise \\
                \end{array} \right.
      \label{eq:act_sign}
    \end{equation}
    
    During the binarization procedure, the information on the model parameters should be reserved as much as possible to ensure that the binary model can reproduce the effect of the full-precision model. Inspired by the binarization procedure for the NLP transformer model \cite{liu2022bit}, it can transfer maximal information from full-precision to binarized by making the original weights zero-mean before feeding them to the binarization function. Next, we binarize the weights and activations using the above-mentioned binarization functions:
    
    \begin{equation}
            W_b = \alpha \ast Bi(W_r-\bar{W_r}), ~~ \alpha = \frac{1}{n_W}||W_r||_{l1}
        \label{eq:bi_w}
    \end{equation}
    and
    \begin{equation}
            A_b = \beta \ast Bi(A_r), ~~ \beta = \frac{1}{n_A}||A_r||_{l1},
        \label{eq:bi_act}
    \end{equation}
    where $\alpha$ and $\beta$ are scale factors used to adjust the amplitude of the linear layer outputs, which play the role to prevent the occurrence of the huge distortion compared with the full-precision results. The subscript $b$ indicates binarized values and the subscript $r$ indicates full-precision real values. 
    Then linear layer outputs can be calculated by bitwise operations:
    \begin{equation}
        \begin{aligned}
            Y &= W_b * A_b \\ &= \alpha \ast \beta \ast (Bi(W_r-\bar{W_r}) \circledcirc (Bi(A_r)))
        \end{aligned}
        \label{eq:bi_linear_out}
    \end{equation}
    where the symbol $\circledcirc$ denotes matrix multiplication implemented by the bitwise operators $xnor$ and $bitcount$, whose operation rules are shown in Fig. \ref{fig:bit_op}. In this way, model inference can be significantly accelerated.

    \subsection{Binary transformer module}
    \label{subsec:bi_trans_module}
    
    The details of our binary transformer model are shown in Fig.~\ref{fig:transformer_model}. There are five sub-figures in this figure, from left to right, which are the whole point cloud transformer model, full-precision transformer block, full-precision transformer module, binary transformer module, and binary transformer block respectively. (In this paper, the transformer block is regarded as a part of the transformer module.)
    
    For the convenience of description, the point cloud transformer model is divided into three main modules in Fig.~\ref{fig:pct_a}. Apart from the neighbor feature module with a small graph neural network and variable task-related classifier module, the transformer module with a large number of parameters is the focus of our attention in this paper. 
    The structure of the full-precision transformer module is illustrated briefly in section~\ref{subsec:fp-pct} and also shown in \cref{fig:pct_b,fig:pct_c}. The corresponding binary transformer module is shown in Fig.~\ref{fig:pct_d} and the internal details of whose transformer block is shown in Fig.~\ref{fig:pct_e}. To illustrate the distinctions between full-precision and binary transformer modules, we utilize different colors to annotate network layers, which is orange words and boxes indicate the same structures both in the full-precision and the binary transformer model, while Red words and boxes indicate the different network layers and activations between the full-precision and the binary transformer model. 
    
    Specifically, we divide the transformer module into three components called the input linear block, transformer block, and output linear block respectively as mentioned in section~\ref{subsec:fp-pct}. 
    The symbol $\rightarrow$ denotes the binarization process for weights, red boxes denote binarized activations, and orange boxes denote the output of matrix multiplication (or $xnor$ and $bitcount$ for binary values) between weights and activations. In a word, we binarized all of the activations and weights in the transformer module. In this way, the pipeline of the original network is preserved, so without redesigning the network, binarization can compress the model and accelerate model inference rapidly.

    \subsection{Feature Aggregation and Metric Learning}
	\label{subsec:aggregation}
	In the classification task, the full-precision classifier header is reserved to produce class labels, which is the same setting as the full-precision transformer model. As for the place recognition task, we find the NetVLAD \cite{NetVlad} layer often used in the previous works, such as \cite{Liu-LPD-Net,Uy2018PointNetVLADDP} is not necessary for our work. So the NetVLAD layer is replaced by a simple max-pooling layer for aggregating a globally consistent descriptor after the last linear layers for class labels generation are removed.

\begin{figure}[t]
      \centering
       \includegraphics[width=0.5\linewidth]{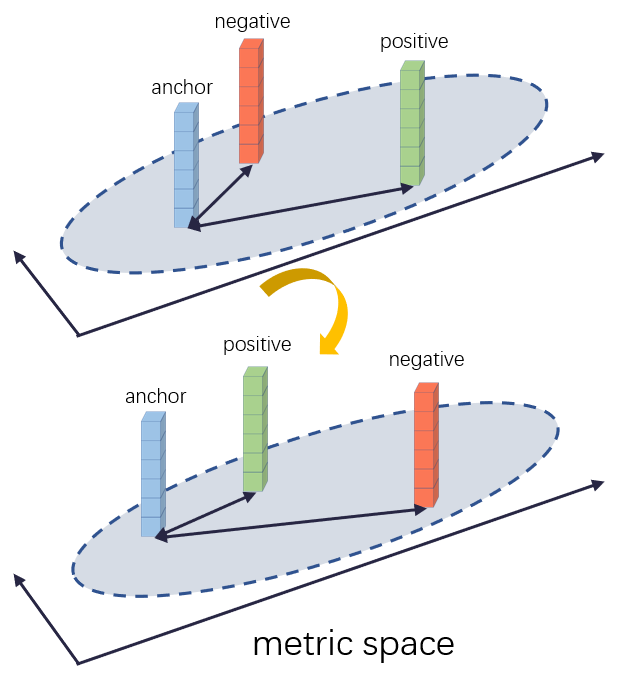}
       \caption{Loss function modifies the relative distances in the metric space.}
       \label{fig:metric_loss}
    \end{figure}

    In respect of the loss function, the naive soft cross-entropy loss is used for point cloud classification while the metric learning loss is required for the place recognition task shown in Fig. \ref{fig:metric_loss}. Specifically, the loss function we used is a role for minimizing the relative distance between descriptors extracted from point clouds sampled at the revisited place (defined as positive samples) and maximizing the relative distance between descriptors extracted from point clouds sampled at different places (defined as negative samples) in the metric space. The lazy quadruplet loss proposed in \cite{Uy2018PointNetVLADDP} used in our work is defined as Eq.\ref{equ:lazy-quadruplet}. In this loss, the quadruplet consists of a query sample, a positive sample set, a negative sample set, and another negative sample, constructed as $T=(P_a, \{P_p\}, \{P_n\}, P_n^*)$.

    \begin{footnotesize}
    	\begin{equation}
    	\label{equ:lazy-quadruplet}
        	\begin{aligned}		
            	L &= \max \limits_{\substack{i=1...N_p\\j=1...N_n}}([\|f(P_a)-f(P_p^i)\|_2^2 - \|f(P_a)-f(P_n^j)\|_2^2 + \gamma]_+) \\
            	&+\max \limits_{\substack{i=1...N_p\\j=1...N_n}}
            	([\|f(P_a)-f(P_p^i)\|_2^2 - \|f(P_n^*)-f(P_n^j)\|_2^2 + \theta]_+),
        	\end{aligned}
    	\end{equation}	
    \end{footnotesize}
    where $P_p^i \in \{P_p\} $ and $P_n^j \in \{P_n\}$, denote a collection of positive samples and a collection of negative samples respectively, $N_p$ and $N_n$ represent the number of positive samples and negative samples, respectively, in the quadruplet, and $f$ denotes the network to produce the global descriptor. In addition, $\gamma$ and $\theta$ are two constant hyper-parameters to balance two constraint terms.
 
	\section{Experiments}
	\label{sec:result}
    In this section, we validate the effectiveness of the proposed BPT model on the point cloud classification task. And then we apply the BPT model to a larger dataset to handle the place recognition task. This content is divided into three parts. First, we introduce the implementation details and the dataset we used (Sec.~\ref{subsec:exp_settings}). Second, we demonstrate the detailed results on the open benchmark (Sec.~\ref{subsec:main_result}). Finally, we compare the consumption of computing resources by the models with different precision parameters (Sec.~\ref{subsec:com_cost}).
 
    \subsection{Experimental settings}
    \label{subsec:exp_settings}
    
    \noindent\textbf{Classification dataset}: We validate the effectiveness of the proposed BPT model on the point cloud classification benchmark ModelNet40~\cite{wu20153d} dataset. This dataset is the most popular benchmark for point cloud classification. A total of 12,311 CAD models from the 40 categories are split into 9,843 for training and 2,468 for testing.
    
    \noindent\textbf{Place recognition datasets:} We use the benchmark datasets created in \cite{Uy2018PointNetVLADDP} to train the BPT model for the place recognition task. This frequently-used benchmark contains a modified Oxford RobotCar dataset \cite{RobotCarDatasetIJRR} for training and testing, as well as three smaller in-house datasets: University Sector(U.S.), Residential Area(R.A.), and Business District(B.D.) only for testing. The dataset settings are also the same as the ones in \cite{Uy2018PointNetVLADDP}. 

    \noindent\textbf{Evaluation metrics:} The same as the previous works, overall accuracy (OA) and mean accuracy (mAcc) are used to evaluate the classification performance of the proposed model. While the top 1\% (@1\%) and top 1 (@1) average recall rate as the evaluation metrics are adopted to evaluate the place recognition performance of the algorithm.
 
    \noindent\textbf{Classification implementation details:} Our BPT model is based on the full-precision PCT model~\cite{guo2021pct}. So a part of experimental settings including the pre-processing for the input data, the number of network layers, and layer channels all follow this work. The proposed BPT model is trained from scratch without leveraging any pre-trained model. We use the stochastic gradient descent (SGD) optimizer and an initial learning rate of 0.01 with 0.9 momentum to train our network 500 epochs iteratively on the Pytorch platform. A cosine annealing schedule is adopted to adjust the learning rate. The mini-batch size is set to 32. All the evaluations are conducted on an NVIDIA RTX2080Ti GPU card.
    
	\begin{table}[t]
    	\caption{Comparison results for point cloud classification.} 
		\label{tab:result_tmp}
		\centering
            \rowcolors{2}{gray!50}{gray!20}
		\begin{tabular}{c|c|c|c|c}  
			\hline       
			\hline
			\rowcolor{red!30}Methods         &Transformer& W-A      &  OA(\%)   & mAcc(\%)      \\
			\hline
                \hline
			PointNet \cite{pointnet}        &\ding{55}& 32-32    & 89.2 & 86.0     \\
			\hline
   		BiPointNet \cite{qin2020bipointnet} &\ding{55}&  1-1  & 86.4 &  -        \\
			\hline
			PointNet++ \cite{Qi2017PointNetDH} &\ding{55}& 32-32 & 90.7 &  -          \\
			\hline
			DGCNN \cite{Wang2019DynamicGC}    &\ding{55} & 32-32 & 92.9 & 90.0       \\
			\hline
			PCT \cite{guo2021pct}             &\ding{52} & 32-32 & 93.2 & 90.2    \\
			\hline
			\bf{BPT(Ours)}                    &\ding{52} &  1-1  & \textbf{93.4} & \textbf{90.5}  \\ 
			\hline
			\hline
		\end{tabular}
	\end{table}	
     
	\noindent\textbf{Place recognition implementation details:} The network structure for the place recognition follows the structure used for classification except for the classifier header. The dimension of the global descriptor produced by the last pooling layer is set to 256. We feed the final descriptors into the lazy quadruplet loss function and set the hyper-parameters $\gamma$ and $\theta$ in the loss function to 0.5 and 0.2, respectively. We wrap 1 anchor sample, 2 positive samples, and 8 negative samples together as a mini-batch and synchronously input them into two NVIDIA TITAN RTX GPU cards. The optimizer Adam \cite{kingma2014adam} with an initial learning rate of $5\times10^{-5}$ is used to train our network 20 epochs iteratively on the Pytorch platform. The learning rate decays to $1\times10^{-5}$ eventually. 

	\begin{table*}[t]
    	\caption{Comparison results on the average recall at top 1\% and top 1 of different baseline networks trained on Ox. and tested on Ox., U.S., R.A. and B.D., respectively.} 
		\label{tab:baseline-result}
		\centering
		\rowcolors{4}{gray!50}{gray!20}
		\begin{tabular}{c|c|c|c|c|c|c|c|c|c|c}  
			\hline       
			\hline
			\rowcolor{red!30}
            && & \multicolumn{4}{c|}{Ave recall @ 1\%} & \multicolumn{4}{c}{Ave recall @ 1}\\    
			\cline{4-11} \rowcolor{red!30}
			\multirow{-2}{15em}{\centering Methods}&\multirow{-2}{5em}{\centering Transformer}&\multirow{-2}{4em}{\centering Binary Operation}& Ox. & U.S. & R.A. & B.D. & Ox. & U.S. & R.A. & B.D. \\
			\hline
            \hline
			PN\_VLAD \cite{Uy2018PointNetVLADDP} &\ding{55} &\ding{55} &80.33&72.63&60.27&65.30&62.76&65.96&55.31&58.87\\
			\hline
			PCAN \cite{Zhang2019PCAN3A} &\ding{55}&\ding{55}&83.81&79.05&71.18&66.82&69.05&62.50&57.00&58.14\\
			\hline
			DH3D-4096 \cite{du2020dh3d} &\ding{55}&\ding{55}&84.26&-&-&-&73.28&-&-&-\\
			\hline
			DAGC \cite{DAGC}&\ding{55}&\ding{55}&87.49&83.49&75.68&71.21&73.34&-&-&-\\
			\hline
			LPD-Net* \cite{Liu-LPD-Net}&\ding{55}& \ding{55}&91.61&86.02&78.85&75.36&82.41&77.25&65.66&69.51 \\

			\hline
			
			HiTPR \cite{hou2022hitpr}&\ding{52}&\ding{55} &93.71&90.21&87.16&79.79&86.63&80.86&78.16&74.26 \\ 
   		\hline
			EPC-Net \cite{Hui2021Efficient3P}&\ding{55}&\ding{55} &94.74&96.52&88.58&84.92&86.23&-&-&- \\
			\hline
   		PCT \cite{guo2021pct} &\ding{52}&\ding{55}                     
                &93.26&89.72&87.11&78.81&85.98&80.43&77.56&74.01 \\ 
			\hline
            BPT (ours) &\ding{52}&\ding{52} &93.28&89.34&86.55&78.47&85.74&80.46&77.37&74.11 \\ 
            \hline
            \hline
        \end{tabular}
	\end{table*}
 
	\subsection{Quantitative Results}
	\label{subsec:main_result}
    \noindent\textbf{Classification results: }Experimental results on point cloud classification are shown in Tab. \ref{tab:result_tmp}. The bit number of the weights and activations is marked in the W-A column. In this table, PCT means full-precision point cloud transformer which is the counterpart of our binarized model. We compare our binary point cloud transformer with three kinds of methods, one is the full-precision method without transformer mechanism, such as PointNet\cite{pointnet}, DGCNN\cite{Wang2019DynamicGC}, etc., one is the full-precision transformer method (PCT\cite{guo2021pct}), and the last is the binary model (BiPointNet\cite{qin2020bipointnet}) binarized from PointNet \cite{pointnet}. This experiment demonstrates the effectiveness of the proposed binary point cloud transformer model whose classification results outperform the compared methods and even surpass its full-precision counterpart.
    
    \noindent\textbf{Place recognition results: }The results of the compared methods are shown in Tab. \ref{tab:baseline-result}. It demonstrates our proposed binary point cloud transformer model (BPT) can achieve competitive results with the existing full-precision models and even surpass the full-precision ones on some indicators with smaller memory space and lighter calculation load. 
    It should be noted that the LPD-Net* indicates the handcrafted features are removed from the original LPD-Net \cite{Liu-LPD-Net} for a fair comparison. We remove the handcrafted features based on the fact that most previous works in LiDAR-based place recognition do not concatenate additional handcrafted features as their network input. 
	
    \subsection{Computational Cost Analysis}
    \label{subsec:com_cost}
\begin{figure}[t]
		\centering
		\subfigure[model size (56.1\%)]{
			\begin{minipage}[t]{0.2\textwidth}
				\centering
				\includegraphics[width=1\textwidth]{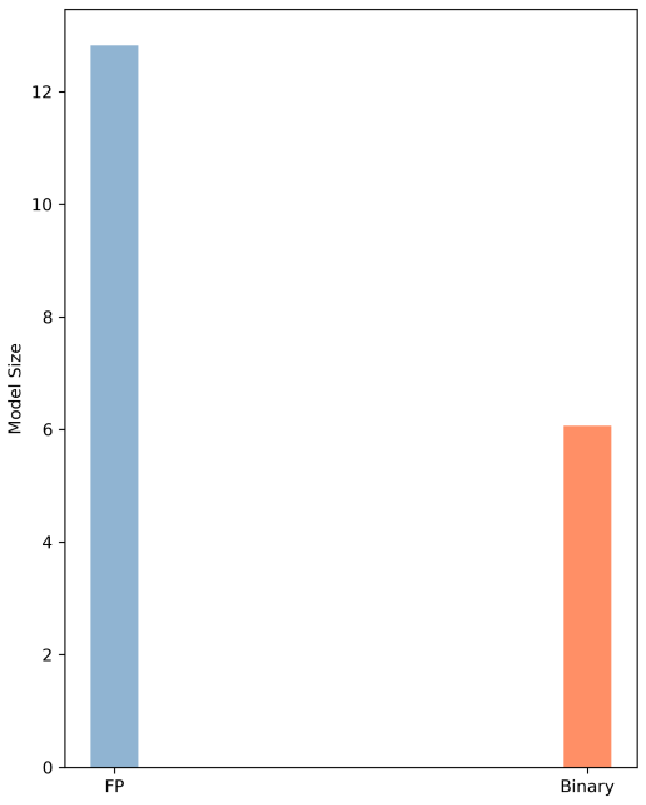}
			\end{minipage}%
		}%
		\subfigure[FLOPs (34.1\%)]{
			\begin{minipage}[t]{0.2\textwidth}
				\centering
				\includegraphics[width=1\textwidth]{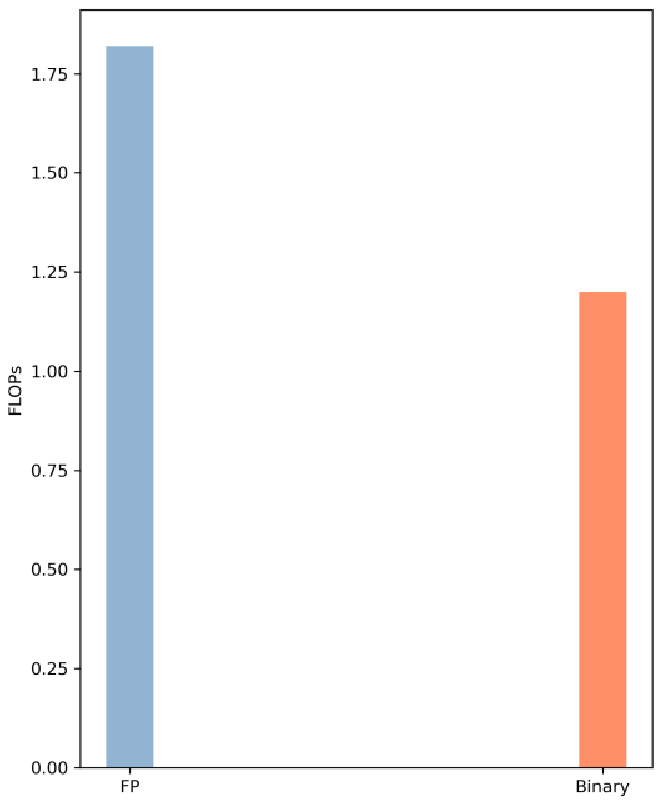}
			\end{minipage}%
		}%
		\centering
		\caption{Calculation resource consumption comparison. FP means the full-precision model, while Binary means the proposed binary point cloud transformer model.}
		\label{fig:resource}
	\end{figure}

        We analyze the consumption of computing resources from two aspects: the model size and FLOPs (floating point operations). After binarization, the size of the same model structure reduces 56.1\% from 12.9M to 5.7M, while the FLOPs reduce 34.1\% from 1.82G to 1.20G shown in Fig. \ref{fig:resource}. Compared with the full-precision point cloud transformer, the proposed binary one is more suitable to be deployed on a mobile robot in real-time application scenarios, such as the SLAM system.
	
    \section{Conclusion}
    \label{sec:conclusion}
        In this paper, we propose a binary point cloud transformer network to extract a discriminative and generalized descriptor from the point cloud for place recognition, which compresses model parameters and activations from 32-bit full-precision to 1-bit binarized, and results in less memory occupation and faster inference speed. To verify the effectiveness of the proposed model, a simple point cloud classification experiment and a place recognition experiment in four different scenarios are conducted. Experiments demonstrate the proposed binary point transformer obtains competitive performance and even surpasses the existing full-precision model in some indicators. Furthermore, the proposed BPT model has the advantage of small memory occupation and fast inference speed so it is extremely suitable for actual deployment to mobile robots.



    \bibliographystyle{IEEEtran}
    \bibliography{example}  

\begin{thebibliography}{10}
\providecommand{\url}[1]{#1}
\csname url@rmstyle\endcsname
\providecommand{\newblock}{\relax}
\providecommand{\bibinfo}[2]{#2}
\providecommand\BIBentrySTDinterwordspacing{\spaceskip=0pt\relax}
\providecommand\BIBentryALTinterwordstretchfactor{4}
\providecommand\BIBentryALTinterwordspacing{\spaceskip=\fontdimen2\font plus
\BIBentryALTinterwordstretchfactor\fontdimen3\font minus
  \fontdimen4\font\relax}
\providecommand\BIBforeignlanguage[2]{{%
\expandafter\ifx\csname l@#1\endcsname\relax
\typeout{** WARNING: IEEEtran.bst: No hyphenation pattern has been}%
\typeout{** loaded for the language `#1'. Using the pattern for}%
\typeout{** the default language instead.}%
\else
\language=\csname l@#1\endcsname
\fi
#2}}

\bibitem{Wang2020LiDARIF}
Y.~Wang, Z.~Sun, C.~Xu, S.~Sarma, J.~Yang, and H.~Kong, ``Lidar iris for
  loop-closure detection,'' \emph{IROS}, pp. 5769--5775, 2020.

\bibitem{kim2018scan}
G.~Kim and A.~Kim, ``Scan context: Egocentric spatial descriptor for place
  recognition within 3d point cloud map,'' in \emph{IROS}, 2018, pp.
  4802--4809.

\bibitem{segmatch2017}
R.~Dub{\'e}, D.~Dugas, E.~Stumm, J.~Nieto, R.~Siegwart, and C.~Cadena,
  ``Segmatch: Segment based place recognition in 3d point clouds,'' in
  \emph{ICRA}, 2017, pp. 5266--5272.

\bibitem{Uy2018PointNetVLADDP}
M.~A. Uy and G.~H. Lee, ``Pointnetvlad: Deep point cloud based retrieval for
  large-scale place recognition,'' \emph{CVPR}, pp. 4470--4479, 2018.

\bibitem{Liu-LPD-Net}
Z.~Liu, S.~Zhou, C.~Suo, P.~Yin, W.~Chen, H.~Wang, H.~Li, and Y.~Liu,
  ``Lpd-net: 3d point cloud learning for large-scale place recognition and
  environment analysis,'' in \emph{ICCV}, 2019, pp. 2831--2840.

\bibitem{du2020dh3d}
J.~Du, R.~Wang, and D.~Cremers, ``Dh3d: Deep hierarchical 3d descriptors for
  robust large-scale 6dof relocalization,'' in \emph{ECCV}, 2020.

\bibitem{vid2021locus}
K.~Vidanapathirana, P.~Moghadam, B.~Harwood, M.~Zhao, S.~Sridharan, and
  C.~Fookes, ``Locus: Lidar-based place recognition using spatiotemporal
  higher-order pooling,'' in \emph{ICRA}, 2021.

\bibitem{Xia2020SOENetAS}
Y.~Xia, Y.~Xu, S.~Li, R.~Wang, J.~Du, D.~Cremers, and U.~Stilla, ``Soe-net: A
  self-attention and orientation encoding network for point cloud based place
  recognition,'' \emph{CVPR}, 2021.

\bibitem{Zhou2021NDTTransformerL3}
Z.~Zhou, C.~Zhao, D.~Adolfsson, S.~Su, Y.~Gao, T.~Duckett, and L.~Sun,
  ``Ndt-transformer: Large-scale 3d point cloud localisation using the normal
  distribution transform representation,'' \emph{ICRA}, 2021.

\bibitem{hou2022hitpr}
Z.~Hou, Y.~Yan, C.~Xu, and H.~Kong, ``Hitpr: Hierarchical transformer for place
  recognition in point cloud,'' in \emph{2022 International Conference on
  Robotics and Automation (ICRA)}, 2022, pp. 2612--2618.

\bibitem{lcdnet}
D.~Cattaneo, M.~Vaghi, and A.~Valada, ``Lcdnet: Deep loop closure detection for
  lidar slam based on unbalanced optimal transport,''
  \emph{arXiv:2103.05056v1}, 2021.

\bibitem{Komorowski_2021_WACV}
J.~Komorowski, ``Minkloc3d: Point cloud based large-scale place recognition,''
  in \emph{WACV}, January 2021, pp. 1790--1799.

\bibitem{chen2020overlapnet}
X.~Chen, T.~L{\"a}be, A.~Milioto, T.~R{\"o}hling, O.~Vysotska, A.~Haag,
  J.~Behley, C.~Stachniss, and F.~Fraunhofer, ``Overlapnet: Loop closing for
  lidar-based slam,'' in \emph{Proc. of Robotics: Science and Systems (RSS)},
  2020.

\bibitem{he2016m2dp}
L.~He, X.~Wang, and H.~Zhang, ``M2dp: A novel 3d point cloud descriptor and its
  application in loop closure detection,'' in \emph{IROS}, 2016, pp. 231--237.

\bibitem{devlin2019bert}
J.~Devlin, M.~Chang, K.~Lee, and K.~Toutanova, ``Bert: Pre-training of deep
  bidirectional transformers for language understanding,'' in \emph{NAACL-HLT},
  2019, pp. 4171--4186.

\bibitem{brown2020language}
T.~Brown, B.~Mann, N.~Ryder, M.~Subbiah, J.~D. Kaplan, P.~Dhariwal,
  A.~Neelakantan, P.~Shyam, G.~Sastry, A.~Askell, S.~Agarwal, A.~Herbert-Voss,
  G.~Krueger, T.~Henighan, R.~Child, A.~Ramesh, D.~Ziegler, J.~Wu, C.~Winter,
  C.~Hesse, M.~Chen, E.~Sigler, M.~Litwin, S.~Gray, B.~Chess, J.~Clark,
  C.~Berner, S.~McCandlish, A.~Radford, I.~Sutskever, and D.~Amodei, ``Language
  models are few-shot learners,'' in \emph{NeurIPS}, vol.~33, 2020, pp.
  1877--1901.

\bibitem{dosovitskiy2020vit}
A.~Dosovitskiy, L.~Beyer, A.~Kolesnikov, D.~Weissenborn, X.~Zhai,
  T.~Unterthiner, M.~Dehghani, M.~Minderer, G.~Heigold, S.~Gelly, J.~Uszkoreit,
  and N.~Houlsby, ``An image is worth 16x16 words: Transformers for image
  recognition at scale,'' \emph{ICLR}, 2021.

\bibitem{liu2021swin}
Z.~Liu, Y.~Lin, Y.~Cao, H.~Hu, Y.~Wei, Z.~Zhang, S.~Lin, and B.~Guo, ``Swin
  transformer: Hierarchical vision transformer using shifted windows,'' in
  \emph{Proceedings of the IEEE/CVF International Conference on Computer
  Vision}, 2021, pp. 10\,012--10\,022.

\bibitem{zhao2021point}
H.~Zhao, L.~Jiang, J.~Jia, P.~H. Torr, and V.~Koltun, ``Point transformer,'' in
  \emph{Proceedings of the IEEE/CVF International Conference on Computer
  Vision}, 2021, pp. 16\,259--16\,268.

\bibitem{guo2021pct}
M.-H. Guo, J.-X. Cai, Z.-N. Liu, T.-J. Mu, R.~R. Martin, and S.-M. Hu, ``Pct:
  Point cloud transformer,'' \emph{Computational Visual Media}, vol.~7, no.~2,
  pp. 187--199, 2021.

\bibitem{hui2021pyramid}
L.~Hui, H.~Yang, M.~Cheng, J.~Xie, and J.~Yang, ``Pyramid point cloud
  transformer for large-scale place recognition,'' in \emph{Proceedings of the
  IEEE/CVF International Conference on Computer Vision}, 2021, pp. 6098--6107.

\bibitem{fan2022svt}
Z.~Fan, Z.~Song, H.~Liu, Z.~Lu, J.~He, and X.~Du, ``Svt-net: Super light-weight
  sparse voxel transformer for large scale place recognition,'' in
  \emph{Proceedings of the AAAI Conference on Artificial Intelligence},
  vol.~36, no.~1, 2022, pp. 551--560.

\bibitem{fan2019reducing}
A.~Fan, E.~Grave, and A.~Joulin, ``Reducing transformer depth on demand with
  structured dropout,'' in \emph{International Conference on Learning
  Representations}, 2019.

\bibitem{michel2019sixteen}
P.~Michel, O.~Levy, and G.~Neubig, ``Are sixteen heads really better than
  one?'' \emph{Advances in neural information processing systems}, vol.~32,
  2019.

\bibitem{voita2019analyzing}
E.~Voita, D.~Talbot, F.~Moiseev, R.~Sennrich, and I.~Titov, ``Analyzing
  multi-head self-attention: Specialized heads do the heavy lifting, the rest
  can be pruned,'' in \emph{Proceedings of the 57th Annual Meeting of the
  Association for Computational Linguistics}, 2019, pp. 5797--5808.

\bibitem{jiao2020tinybert}
X.~Jiao, Y.~Yin, L.~Shang, X.~Jiang, X.~Chen, L.~Li, F.~Wang, and Q.~Liu,
  ``Tinybert: Distilling bert for natural language understanding,'' in
  \emph{Findings of the Association for Computational Linguistics: EMNLP 2020},
  2020, pp. 4163--4174.

\bibitem{sun2019patient}
S.~Sun, Y.~Cheng, Z.~Gan, and J.~Liu, ``Patient knowledge distillation for bert
  model compression,'' in \emph{EMNLP/IJCNLP (1)}, 2019.

\bibitem{sanh2019distilbert}
V.~Sanh, L.~Debut, J.~Chaumond, and T.~Wolf, ``Distilbert, a distilled version
  of bert: smaller, faster, cheaper and lighter,'' \emph{arXiv preprint
  arXiv:1910.01108}, 2019.

\bibitem{ma2019tensorized}
X.~Ma, P.~Zhang, S.~Zhang, N.~Duan, Y.~Hou, M.~Zhou, and D.~Song, ``A
  tensorized transformer for language modeling,'' \emph{Advances in neural
  information processing systems}, vol.~32, 2019.

\bibitem{lan2019albert}
Z.~Lan, M.~Chen, S.~Goodman, K.~Gimpel, P.~Sharma, and R.~Soricut, ``Albert: A
  lite bert for self-supervised learning of language representations,'' in
  \emph{International Conference on Learning Representations}, 2019.

\bibitem{hou2020dynabert}
L.~Hou, Z.~Huang, L.~Shang, X.~Jiang, X.~Chen, and Q.~Liu, ``Dynabert: Dynamic
  bert with adaptive width and depth,'' \emph{Advances in Neural Information
  Processing Systems}, vol.~33, pp. 9782--9793, 2020.

\bibitem{liu2020fastbert}
W.~Liu, P.~Zhou, Z.~Wang, Z.~Zhao, H.~Deng, and Q.~Ju, ``Fastbert: a
  self-distilling bert with adaptive inference time,'' in \emph{Proceedings of
  the 58th Annual Meeting of the Association for Computational Linguistics},
  2020, pp. 6035--6044.

\bibitem{zafrir2019q8bert}
O.~Zafrir, G.~Boudoukh, P.~Izsak, and M.~Wasserblat, ``Q8bert: Quantized 8bit
  bert,'' in \emph{2019 Fifth Workshop on Energy Efficient Machine Learning and
  Cognitive Computing-NeurIPS Edition (EMC2-NIPS)}.\hskip 1em plus 0.5em minus
  0.4em\relax IEEE, 2019, pp. 36--39.

\bibitem{zhang2020ternarybert}
W.~Zhang, L.~Hou, Y.~Yin, L.~Shang, X.~Chen, X.~Jiang, and Q.~Liu,
  ``Ternarybert: Distillation-aware ultra-low bit bert,'' in \emph{Proceedings
  of the 2020 Conference on Empirical Methods in Natural Language Processing
  (EMNLP)}, 2020, pp. 509--521.

\bibitem{shen2020q}
S.~Shen, Z.~Dong, J.~Ye, L.~Ma, Z.~Yao, A.~Gholami, M.~W. Mahoney, and
  K.~Keutzer, ``Q-bert: Hessian based ultra low precision quantization of
  bert,'' in \emph{Proceedings of the AAAI Conference on Artificial
  Intelligence}, vol.~34, 2020, pp. 8815--8821.

\bibitem{lizhexin2022qvit}
Z.~Li, T.~Yang, P.~Wang, and J.~Cheng, ``Q-vit: Fully differentiable
  quantization for vision transformer,'' \emph{arXiv preprint
  arXiv:2201.07703}, 2022.

\bibitem{liyanjing2022qvit}
Y.~Li, S.~Xu, B.~Zhang, X.~Cao, P.~Gao, and G.~Guo, ``Q-vit: Accurate and fully
  quantized low-bit vision transformer,'' in \emph{NIPS}, 2022.

\bibitem{bai2021binarybert}
H.~Bai, W.~Zhang, L.~Hou, L.~Shang, J.~Jin, X.~Jiang, Q.~Liu, M.~R. Lyu, and
  I.~King, ``Binarybert: Pushing the limit of bert quantization,'' in
  \emph{ACL/IJCNLP (1)}, 2021.

\bibitem{qin2021bibert}
H.~Qin, Y.~Ding, M.~Zhang, Y.~Qinghua, A.~Liu, Q.~Dang, Z.~Liu, and X.~Liu,
  ``Bibert: Accurate fully binarized bert,'' in \emph{International Conference
  on Learning Representations}, 2021.

\bibitem{liu2022bit}
Z.~Liu, B.~Oguz, A.~Pappu, L.~Xiao, S.~Yih, M.~Li, R.~Krishnamoorthi, and
  Y.~Mehdad, ``Bit: Robustly binarized multi-distilled transformer,''
  \emph{arXiv preprint arXiv:2205.13016}, 2022.

\bibitem{rastegari2016xnor}
M.~Rastegari, V.~Ordonez, J.~Redmon, and A.~Farhadi, ``Xnor-net: Imagenet
  classification using binary convolutional neural networks,'' in
  \emph{European conference on computer vision}.\hskip 1em plus 0.5em minus
  0.4em\relax Springer, 2016, pp. 525--542.

\bibitem{wu2022point}
X.~Wu, Y.~Lao, L.~Jiang, X.~Liu, and H.~Zhao, ``Point transformer v2: Grouped
  vector attention and partition-based pooling,'' \emph{arXiv preprint
  arXiv:2210.05666}, 2022.

\bibitem{wu20153d}
Z.~Wu, S.~Song, A.~Khosla, F.~Yu, L.~Zhang, X.~Tang, and J.~Xiao, ``3d
  shapenets: A deep representation for volumetric shapes,'' in
  \emph{Proceedings of the IEEE conference on computer vision and pattern
  recognition}, 2015, pp. 1912--1920.

\bibitem{ESF}
W.~Wohlkinger and M.~Vincze, ``Ensemble of shape functions for 3d object
  classification,'' in \emph{ROBIO}, 2011, pp. 2987--2992.

\bibitem{point-hist}
R.~B. Rusu, N.~Blodow, Z.~C. Marton, and M.~Beetz, ``Aligning point cloud views
  using persistent feature histograms,'' in \emph{IROS}, 2008, pp. 3384--3391.

\bibitem{fpfh}
R.~B. Rusu, N.~Blodow, and M.~Beetz, ``Fast point feature histograms (fpfh) for
  3d registration,'' in \emph{ICRA}, 2009, pp. 3212--3217.

\bibitem{pointnet}
R.~Q. {Charles}, H.~{Su}, M.~{Kaichun}, and L.~J. {Guibas}, ``Pointnet: Deep
  learning on point sets for 3d classification and segmentation,'' in
  \emph{CVPR}, 2017, pp. 77--85.

\bibitem{NetVlad}
R.~{Arandjelovic}, P.~{Gronat}, A.~{Torii}, T.~{Pajdla}, and J.~{Sivic},
  ``Netvlad: Cnn architecture for weakly supervised place recognition,'' in
  \emph{CVPR}, 2016, pp. 5297--5307.

\bibitem{Wang2019DynamicGC}
Y.~Wang, Y.~Sun, Z.~Liu, S.~E. Sarma, M.~Bronstein, and J.~Solomon, ``Dynamic
  graph cnn for learning on point clouds,'' \emph{ACM Transactions on Graphics
  (TOG)}, vol.~38, pp. 1 -- 12, 2019.

\bibitem{Zhang2019PCAN3A}
W.~Zhang and C.~Xiao, ``Pcan: 3d attention map learning using contextual
  information for point cloud based retrieval,'' \emph{CVPR}, pp.
  12\,428--12\,437, 2019.

\bibitem{Qi2017PointNetDH}
C.~Qi, L.~Yi, H.~Su, and L.~Guibas, ``Pointnet++: Deep hierarchical feature
  learning on point sets in a metric space,'' in \emph{NeurIPS}, 2017.

\bibitem{Hui2021Efficient3P}
L.~Hui, M.~Cheng, J.~Xie, and J.~Yang, ``Efficient 3d point cloud feature
  learning for large-scale place recognition,'' \emph{ArXiv}, vol.
  abs/2101.02374, 2021.

\bibitem{courbariaux2016binarized}
M.~Courbariaux, I.~Hubara, D.~Soudry, R.~El-Yaniv, and Y.~Bengio, ``Binarized
  neural networks: Training deep neural networks with weights and activations
  constrained to+ 1 or-1,'' \emph{arXiv preprint arXiv:1602.02830}, 2016.

\bibitem{vaswani2017attention}
A.~Vaswani, N.~Shazeer, N.~Parmar, J.~Uszkoreit, L.~Jones, A.~N. Gomez,
  {\L}.~Kaiser, and I.~Polosukhin, ``Attention is all you need,'' in
  \emph{NIPS}, vol.~30, 2017.

\bibitem{qin2020bipointnet}
H.~Qin, Z.~Cai, M.~Zhang, Y.~Ding, H.~Zhao, S.~Yi, X.~Liu, and H.~Su,
  ``Bipointnet: Binary neural network for point clouds,'' in
  \emph{International Conference on Learning Representations}, 2020.

\bibitem{xu2021poem}
S.~Xu, Y.~Li, J.~Zhao, B.~Zhang, and G.~Guo, ``Poem: 1-bit point-wise
  operations based on expectation-maximization for efficient point cloud
  processing,'' \emph{arXiv preprint arXiv:2111.13386}, 2021.

\bibitem{su2022svnet}
Z.~Su, M.~Welling, L.~Liu, \emph{et~al.}, ``Svnet: Where so (3) equivariance
  meets binarization on point cloud representation,'' \emph{arXiv preprint
  arXiv:2209.05924}, 2022.

\bibitem{bengio2013estimating}
Y.~Bengio, N.~L{\'e}onard, and A.~Courville, ``Estimating or propagating
  gradients through stochastic neurons for conditional computation,''
  \emph{arXiv preprint arXiv:1308.3432}, 2013.

\bibitem{RobotCarDatasetIJRR}
W.~Maddern, G.~Pascoe, C.~Linegar, and P.~Newman, ``{1 Year, 1000km: The Oxford
  RobotCar Dataset},'' \emph{IJRR}, vol.~36, no.~1, pp. 3--15, 2017.

\bibitem{kingma2014adam}
D.~P. Kingma and J.~Ba, ``Adam: A method for stochastic optimization,'' in
  \emph{ICLR}, 2015.

\bibitem{DAGC}
Q.~Sun, H.~Liu, J.~He, Z.~Fan, and X.~Du, ``Dagc: Employing dual attention and
  graph convolution for point cloud based place recognition,'' in \emph{ICMR},
  2020, p. 224–232.

\end{thebibliography}
	
\end{document}